\documentclass{article}
\usepackage[T1]{fontenc}
\usepackage[utf8]{inputenc}
\usepackage{ismir} 
\usepackage{amsmath,amssymb,mathtools,amsthm}
\usepackage{amsmath,cite,url}
\usepackage{graphicx}
\usepackage{color}

\title{Paper Template for ISMIR \conferenceyear}

\usepackage{microtype}
\usepackage{subcaption}
\usepackage{booktabs}
\usepackage{multirow}
\usepackage{tabularx}
\usepackage{enumitem}
\usepackage{placeins}

\usepackage{tikz}
\usetikzlibrary{arrows.meta,positioning,fit,calc,backgrounds}

\usepackage{rotating}
\usepackage{lscape}

\usepackage[disable,textsize=tiny]{todonotes}

\title{TART: A Modular Tool for Technique-Aware Audio-to-Tablature Guitar Transcription}

\multauthor
  {Akshaj Gupta \hspace{1cm} Hwi Joo Park \hspace{1cm} Andrea Guzman \hspace{1cm} Shamak Gowda}
  {{\bf Samhita Konduri \hspace{1cm} Jiachen Lian \hspace{1cm} Robin Netzorg \hspace{1cm} Gopala Anumanchipalli}\\
  University of California, Berkeley\\
  {\tt\small akshaj.gupta@berkeley.edu}
  }

\def\authorname{A. Gupta, H. J. Park, A. Guzman, S. Gowda, S. Konduri, J. Lian, R. Netzorg, and G. Anumanchipalli}

\usepackage[bookmarks=false,pdfauthor={\authorname},pdfsubject={\pdfsubject},hidelinks]{hyperref}

\begin{document}
\raggedbottom

\maketitle

\begin{abstract}
  Automatic Music Transcription (AMT) for guitar remains limited by three challenges: existing systems often fail to capture expressive techniques such as slides, bends, and percussive hits; they often assign notes to incorrect string-fret combinations; and they are typically trained on clean recordings, limiting generalization to noisy real-world audio. To address these challenges, we propose TART, a modular four-stage audio-to-tablature pipeline consisting of (1) an audio-to-MIDI transcription model, (2) an expressive technique classifier, (3) an audio-conditioned T5 encoder-decoder for string-fret assignment, and (4) an automated tablature generator. We evaluate TART in a zero-shot setting on GuitarSet, EGDB, and two augmented benchmarks, Noisy GuitarSet and Noisy EGDB. Averaged across these four benchmarks, TART achieves \textbf{$81.35\%$ audio-to-MIDI F50} ($+6.67$ points over the best prior baseline), \textbf{$71.8\%$ string-fret Tab F1} ($+8.5$ points over the best prior baseline), and an overall \textbf{54.08\% end-to-end Tab F1}. To our knowledge, TART is the first framework to generate guitar tablature with both fingering and expressive technique annotations directly from guitar audio.

\end{abstract}

\section{Introduction}\label{sec:introduction}

Automatic Music Transcription (AMT) is the process of converting audio recordings into symbolic representations such as sheet music, MIDI, or tablature (tab). Audio-to-tab transcription is a long-standing problem and has proven to be much harder than transcribing piano, where deep learning has driven CRNN-based models to high note-level accuracy~\cite{hawthorne-2018-onsets, kong2021highresolutionpianotranscriptionpedals}. While guitar transcription systems have adopted similar architectures~\cite{riley2024highresolutionguitartranscription, bittner2022lightweightinstrumentagnosticmodelpolyphonic, hamberger2025frettingtransformerencoderdecodermodelmidi, maman2022unalignedsupervisionautomaticmusic}, they still fall short of producing accurate tabs for several reasons.

First, guitarists rely heavily on expressive techniques such as slides, bends, harmonics, and percussive hits, none of which current AMT systems capture. Second, the guitar is pitch-redundant, meaning the same pitch can be played at several different string-fret combinations.  Existing transcription systems often assign incorrect combinations, producing tablature that does not reflect how a guitarist would physically play the piece. Third, most guitar-specific models are trained on small, professionally-recorded datasets and generalize poorly to noisy, real-world recordings where transcription is most useful.

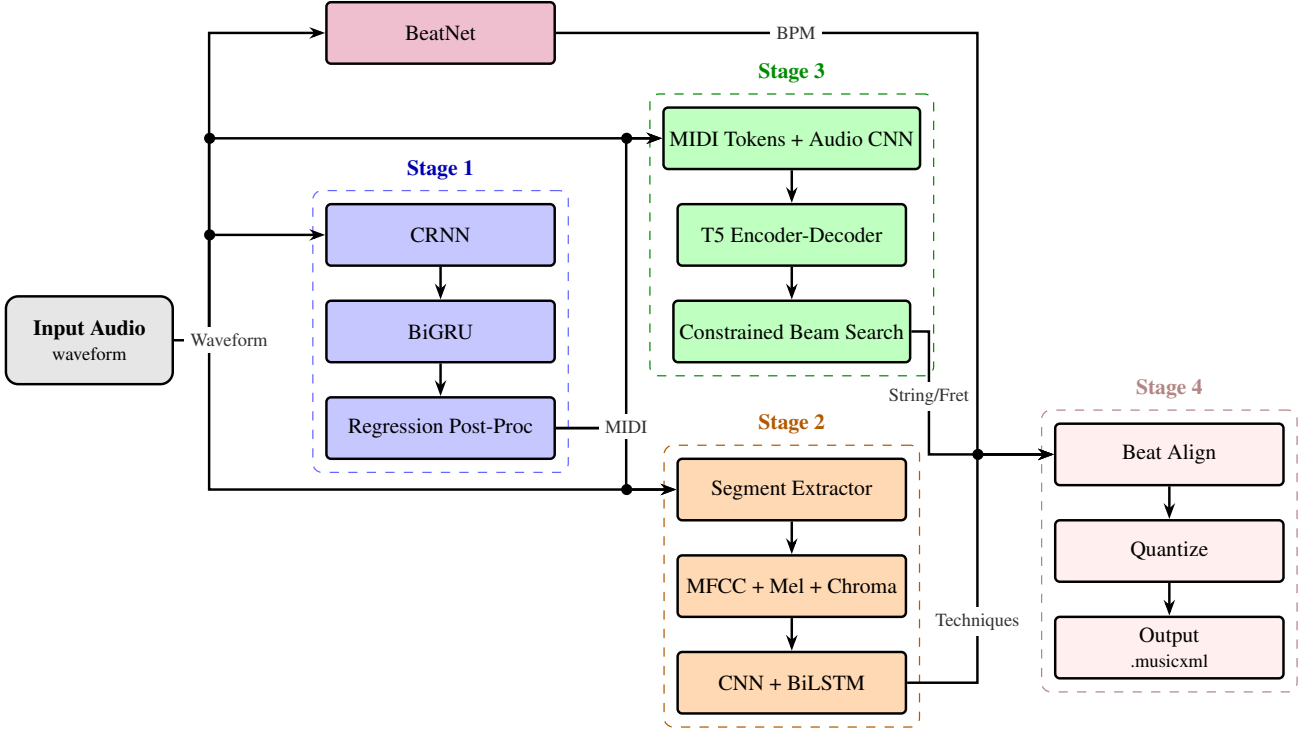
\begin{figure*}
\centering
\resizebox{1.0\textwidth}{!}{%
\begin{tikzpicture}[
    x=1cm, y=1cm,
    sub/.style={rectangle, draw, minimum width=2.6cm, minimum height=0.7cm, align=center, font=\scriptsize, rounded corners=1.5pt, thick, inner sep=2pt},
    beat/.style={sub, fill=purple!25},
    s1/.style={sub, fill=blue!22},
    s2/.style={sub, fill=orange!30},
    s3/.style={sub, fill=green!25},
    s4/.style={sub, fill=pink!25},
    io/.style={sub, fill=gray!20, rounded corners=4pt, font=\scriptsize, minimum width=1.9cm, minimum height=1cm},
    arrow/.style={-{Stealth[length=1.8mm, width=1.3mm]}, thick},
    datalbl/.style={font=\tiny, inner sep=2pt, fill=white, text=black!80},
    stagelbl/.style={font=\scriptsize\bfseries},
    group/.style={draw, dashed, rounded corners=3pt, inner sep=4pt}
]

\node[io] (input) at (0, 0) {\textbf{Input Audio}\\\tiny waveform};

\node[beat] (beatnet) at (4, 3.5) {BeatNet};

\node[s1] (crnn)  at (4, 1.2) {CRNN};
\node[s1] (bigru) at (4, 0.1) {BiGRU};
\node[s1] (regpp) at (4, -1) {Regression Post-Proc};

\node[s3] (tok)  at (8, 2.3) {MIDI Tokens + Audio CNN};
\node[s3] (t5)   at (8, 1.2) {T5 Encoder-Decoder};
\node[s3] (beam) at (8, 0.1) {Constrained Beam Search};

\node[s2] (seg)  at (8, -1.7) {Segment Extractor};
\node[s2] (feat) at (8, -2.8) {MFCC + Mel + Chroma};
\node[s2] (cnn2) at (8, -3.9) {CNN + BiLSTM};

\node[s4] (stage4)  at (12.3, -1.3) {Beat Align};
\node[s4] (quantize) at (12.3, -2.4) {Quantize};
\node[s4] (out) at (12.3, -3.5) {Output \\\tiny .musicxml};

\fill (10.12,-1.3) circle (1.8pt);
\fill (6.12, 2.3) circle (1.8pt);
\fill (6.12, -1.7) circle (1.8pt);
\fill (1.37, 1.2) circle (1.8pt);
\fill (1.37, 2.3) circle (1.8pt);

\begin{scope}[on background layer]
    \node[group, draw=blue!60, fit=(crnn)(bigru)(regpp),
          label={[stagelbl, text=blue!70!black]above:Stage 1}] {};
    \node[group, draw=green!55!black, fit=(tok)(t5)(beam),
          label={[stagelbl, text=green!55!black]above:Stage 3 }] {};
    \node[group, draw=orange!70!black, fit=(seg)(feat)(cnn2),
          label={[stagelbl, text=orange!70!black]above:Stage 2}] {};
    \node[group, draw=pink!70!black, fit=(stage4)(quantize)(out),
          label={[stagelbl, text=pink!70!black]above:Stage 4}] {};
\end{scope}

\path (input.east) ++(0.4, 0) coordinate (abus);

\draw[arrow] (abus) |- (beatnet.west);
\draw[arrow] (abus) |- (crnn.west);
\draw[arrow] (abus) |- (tok.west);
\draw[arrow] (abus) |- (seg.west);

\draw[arrow] (crnn) -- (bigru);
\draw[arrow] (bigru) -- (regpp);
\draw[arrow] (tok) -- (t5);
\draw[arrow] (t5) -- (beam);
\draw[arrow] (seg) -- (feat);
\draw[arrow] (feat) -- (cnn2);
\draw[arrow] (stage4) -- (quantize);
\draw[arrow] (quantize) -- (out);

\draw[arrow] (regpp.east) -- ++(0.8,0) |- (tok.west);
\draw[arrow] (regpp.east) -- ++(0.8,0) |- (seg.west)
    node[datalbl, pos=0.0, above, yshift=-4pt] {MIDI};
\path (regpp.east) ++(0.8, 0) coordinate (mbus);


\draw[arrow] (beatnet.east) -- ++(4.8,0) |- (stage4.west)
    node[datalbl, pos=0.00, left, xshift=-50pt] {BPM};
\draw[arrow] (beam.east) -- ++(0.2,0) |- (stage4.west)
    node[datalbl, pos=0.20, below] {String/Fret};
\draw[arrow] (cnn2.east) -- ++(0.8,0) |- (stage4.west)
    node[datalbl, pos=0.10, above] {Techniques};

\draw[thick] (input.east) -- (abus) node[datalbl, pos=0.0, right, xshift=3.0pt] {Waveform};

\end{tikzpicture}
}
\caption{Flowchart depicting how the four stages are combined to transcribe input audio (left) to output tabs (right).}
\label{fig:finalpipelineflowchart}
\end{figure*}

To address these issues, we propose the \textbf{Technique-Aware Audio-to-Tablature Representation Tool (TART)}, a modular four-stage pipeline that transcribes raw guitar audio into performance-ready tablature (illustrated in Figure~\ref{fig:finalpipelineflowchart}). Stage~1 transcribes the input audio into a time-aligned MIDI note sequence using a high-resolution CRNN. Stage~2 labels each note with one of nine expressive techniques (e.g., slide, bend, harmonic) using a temporal CNN-BiLSTM classifier. Stage~3 resolves the guitar's pitch redundancy problem by assigning each note a string and fret position using a T5 encoder-decoder conditioned on both the symbolic pitch sequence and the original audio. Stage~4 merges the resulting annotations into a beat-aligned MusicXML tablature score. TART achieves new state-of-the-art zero-shot results across multiple benchmarks.

\section{Prior Work}

AMT has already advanced significantly for piano. Hawthorne et al.~\cite{hawthorne-2018-onsets} introduced a dual-objective CRNN that detects onsets and frame-level pitch jointly, and Kong et al.~\cite{kong2021highresolutionpianotranscriptionpedals} extended this with a high-resolution regression-based model that predicts precise onset and offset times. Conversely, guitar transcription has lagged behind in part due to limited paired audio-MIDI data and the polyphonic nature of the instrument. Riley et al.~\cite{riley2024highresolutionguitartranscription} adapted the Kong et al. backbone to acoustic guitar via domain adaptation and achieved strong results. Maman and Bermano~\cite{maman2022unalignedsupervisionautomaticmusic} focused on addressing data scarcity through unaligned supervision. Despite these advances, current systems struggle with electric pickups, distortion, and noisy real-world recordings.

Expressive technique recognition remains challenging due to limited data and inconsistent labels across datasets. Magcil~\cite{Mitsou_magcil}, AGPT~\cite{Stefani_agpt}, IDMT-SMT-Chords~\cite{kehling_IDMT}, and EG-IPT~\cite{fiorini_2025_15205644} have expanded coverage to bends, slides, harmonics, and percussive techniques, but each focuses on a different subset of techniques and uses incompatible label schemes. Existing technique classifiers are typically feedforward models operating on fixed-length feature vectors~\cite{Stefani_ExpressiveGuitar-TechniqueClassification}, which discard the temporal dynamics that distinguish similar techniques. As a result, no prior system has produced a unified taxonomy with reliable performance.

Resolving string-fret ambiguity has historically been framed as an optimal search problem, but modern solutions fall into two machine learning approaches. \emph{Audio-based models} such as TabCNN~\cite{wiggins2019tabcnn} and FretNet~\cite{Cwitkowitz_2023} predict tablature directly from frame-level audio features, but operate without explicit pitch supervision and struggle with cross-dataset generalization. \emph{Symbolic-based models} treat fingering as sequence-to-sequence translation. Edwards et al.~\cite{edwards2024miditotab} used a BART-style model to predict strings from symbolic pitch, deriving frets deterministically, while the Fretting-Transformer~\cite{hamberger2025frettingtransformerencoderdecodermodelmidi} maps MIDI tokens directly to string-fret tokens with a T5-style encoder-decoder. These symbolic approaches produce playable fingerings from clean MIDI but cannot exploit timbral cues that distinguish identical pitches played on different strings.

\section{First Stage: Audio-to-MIDI Conversion}
\label{sec:stage1_audio2midi}

The goal of Stage~1 is to map an audio waveform $x \in \mathbb{R}^{N}$ (16\,kHz) to a set of MIDI note events
\begin{equation}
\label{eq:stage1_output}
\mathcal{E} = \{(p_i, t_i^{\mathrm{on}}, t_i^{\mathrm{off}}, v_i)\}_{i=1}^{|\mathcal{E}|},
\end{equation}
where $p_i$ is pitch, $t_i^{\mathrm{on}}/t_i^{\mathrm{off}}$ are onset and offset times, and $v_i$ is velocity.

\subsection{Architecture}
\label{subsec:stage1_arch}

We adopt the note-only high-resolution CRNN from Kong et al.~\cite{kong2021highresolutionpianotranscriptionpedals} as our backbone, which takes a log-mel spectrogram as input (16\,kHz audio, 100\,fps, 229 mel bins) and predicts four per-frame, per-class maps: onset confidence, offset confidence, frame activation, and velocity. We leave the base architecture unchanged and instead focus our contributions on adapting the training procedure and post-processing for guitar-specific transcription.

\subsection{Training}
\label{subsec:stage1_training}

A major failure of guitar AMT systems trained on clean acoustic audio is poor transfer to electric pickups, distortion, and noisy consumer recording conditions. To reduce this domain shift, we pool four datasets that span a wide range of instrument domains and recording environments: \textbf{GAPS}~\cite{riley2024gapslargediverseclassical}, \textbf{Guitar-TECHS}~\cite{pedroza2025guitartechselectricguitardataset}, \textbf{Fran\c{c}ois Leduc}~\cite{riley2024highresolutionguitartranscription}, and the DI subset of \textbf{GOAT}~\cite{loth2025goatlargedatasetpaired}. For each dataset, we create an 80/20 train/validation split and merge the splits into a unified pool. We fine-tune the CRNN backbone with Adam (weight decay $10^{-4}$), batch size 4, and learning rate $10^{-5}$ on 30\,s segments with a 10\,s hop, decaying the learning rate by a factor of $0.9$ every 10{,}000 iterations. We select the best model on the validation set.

Beyond dataset diversity, we also propose a new data augmentation strategy. Kong et al.'s~\cite{kong2021highresolutionpianotranscriptionpedals} piano augmentation pipeline applies pitch shifting, dynamic-range compression, equalization, reverberation, and additive noise, but several of these are destructive for guitar: reverb smears attack transients, and aggressive EQ removes harmonics critical for pitch tracking. We instead develop a \textbf{stochastic noise augmentor} that preserves onset alignment while simulating realistic recording-condition noise. With probability $p_{\mathrm{aug}}{=}0.5$ per batch, we apply a zero-phase 80\,Hz high-pass filter and additively mix a random subset of $\{\text{white, pink, 60\,Hz hum}\}$ noise at $\mathrm{SNR}\sim\mathcal{U}(25,45)$\,dB, then peak-normalize to 0.9. To integrate the augmentor into training, we take the checkpoint 2{,}000 iterations before the best non-augmented model and train it for 10{,}000 additional iterations with augmentation enabled, decaying the learning rate by $0.9$ every 5{,}000 iterations and selecting the best model by validation score. To mitigate false positives, we also discard any predicted note shorter than $T_{\min}=30$\,ms. Even at extreme playing speeds (15 notes per second), fully articulated guitar notes typically exceed 50\,ms in duration. 


\begin{table}
\centering
\small
\setlength{\tabcolsep}{4pt}
\renewcommand{\arraystretch}{1.1}
\begin{tabular}{l ccccc}
    \toprule
    \textbf{Model} & \textbf{GS} & \textbf{EGDB} & \textbf{GS noi} & \textbf{EGDB noi} & \textbf{Avg} \\
    \midrule
    FretNet~\cite{Cwitkowitz_2023}       & 69.10 & 40.90 & 37.30 & 23.60 & 42.73 \\
    NoteEM~\cite{maman2022unalignedsupervisionautomaticmusic}        & 82.90 & 59.00 & 70.00 & 67.60 & 69.88 \\
    Riley et al.~\cite{riley2024gapslargediverseclassical}  & \textbf{88.10} & 68.90 & 74.20 & 67.50 & 74.68 \\
    \textbf{TART (Ours)}   & 87.40 & \textbf{79.00} & \textbf{82.20} & \textbf{76.80} & \textbf{81.35} \\
    \bottomrule
\end{tabular}
\caption{Zero-shot audio-to-MIDI transcription results (F50) on GuitarSet (GS), EGDB (EGDB), GuitarSet Noisy (GS noi), and EGDB Noisy (EGDB noi).}
\label{tab:final_comparison}
\end{table}

\subsection{Evaluation}
\label{subsec:stage1_eval}

To evaluate generalization to realistic recording conditions, we generate two new benchmarks: \textbf{Noisy GuitarSet} and \textbf{Noisy EGDB}. To generate them, we apply the same corruption used in our stochastic noise augmentor (Section~\ref{subsec:stage1_training}), with the additional inclusion of a room impulse response (IR) convolution. For each clip in the hex-debleeded subset of GuitarSet~\cite{xi2018guitarset} and the DI subset of EGDB~\cite{Chen2022TowardsAT}, we sample an IR from the EchoThief collection~\cite{warren_echothief}, convolve it with the input signal, and truncate the result to the original length to preserve frame-level alignment between audio and labels. The resulting benchmarks simulate diverse environments (e.g., nature, sanctuaries, venues) and low-quality microphones (e.g., phones, laptops).

We report standard precision, recall, and F1 with an onset tolerance of $\pm 50$\,ms and a pitch tolerance of $\pm 50$\,cents, denoted P50, R50, and F50~\cite{raffel2014mir_eval}. A predicted note is counted as correct only if both its onset and pitch match a ground-truth note within these tolerances. All evaluation is performed in a zero-shot setting on GuitarSet, EGDB, and their noisy counterparts.

Table~\ref{tab:final_comparison} reports zero-shot F50 across all four benchmarks. While the Riley et al.\ model narrowly leads on GuitarSet (88.1\% vs.\ 87.4\%), its performance collapses on EGDB, a pattern consistent with prior systems that tune post-processing thresholds to maximize scores on specific datasets at the expense of generalization. TART achieves the best average F50 score of \textbf{81.35\%}, outperforming the next-best system (Riley et al., 74.68\%) by 6.67 points, with particularly large margins on EGDB ($+10.1$ points), Noisy GuitarSet ($+8.0$ points), and Noisy EGDB ($+9.3$ points).

\section{Second Stage: Expressive Technique Classification}
\label{sec:expressive_classification}

Given Stage~1's note events $\mathcal{E} = \{(p_i, t_i^{\mathrm{on}}, t_i^{\mathrm{off}}, v_i)\}_{i=1}^{|\mathcal{E}|}$ and the input audio $x \in \mathbb{R}^{N}$, Stage~2 assigns each note a technique label $\tau_i \in \mathcal{T}$, producing
\begin{equation}
\label{eq:stage2_output}
\mathcal{E}^{\tau} = \{(p_i, t_i^{\mathrm{on}}, t_i^{\mathrm{off}}, v_i, \tau_i)\}_{i=1}^{|\mathcal{E}|}.
\end{equation}
The label space $\mathcal{T}$ consists of the classes listed in Table~\ref{tab:classification_results}: bend, hammer-on/pull-off, harmonics, kick drum, palm muting, picking/no technique, slide, snare drum, and vibrato.

\subsection{Architecture}
\label{ssec:stage2_architecture}

For each note event in $\mathcal{E}$, we extract the audio chunk spanning the note's duration (from $t_i^{\mathrm{on}}$ to $t_i^{\mathrm{off}}$) and compute a feature sequence at a 23\,ms hop (sample rate 22.05\,kHz, FFT window 1024). Each frame stacks 40 MFCCs, 40 log-mel bands, and 12 chroma coefficients into a 92-dimensional vector, z-normalized per feature across time. Sequences are padded or truncated to a length of 128 frames ($\sim$3\,s), yielding a uniform $128 \times 92$ input tensor.

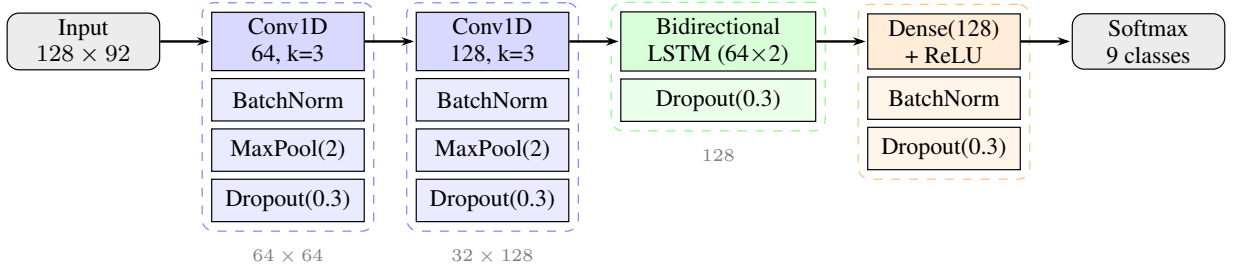
\begin{figure*}[!t]
\centering
\resizebox{0.95\textwidth}{!}{%
\begin{tikzpicture}[
    node distance=0.25cm,
    box/.style={rectangle, draw, minimum width=1.8cm, minimum height=0.5cm, align=center, font=\footnotesize, inner sep=2pt},
    convblock/.style={box, fill=blue!15},
    lstmblock/.style={box, fill=green!15},
    denseblock/.style={box, fill=orange!15},
    ioblock/.style={box, fill=gray!15, rounded corners},
    arrow/.style={-{Stealth[length=1.5mm, width=1.0mm]}, thick, shorten >=0.5pt},
    label/.style={font=\scriptsize, align=center},
    dimlab/.style={font=\tiny, text=gray}
]

\node[ioblock] (input) {Input\\$128 \times 92$};

\node[convblock, right=0.6cm of input] (conv1) {Conv1D\\64, k=3};
\node[box, fill=blue!8, below=0.08cm of conv1] (bn1) {BatchNorm};
\node[box, fill=blue!8, below=0.08cm of bn1] (pool1) {MaxPool(2)};
\node[box, fill=blue!8, below=0.08cm of pool1] (drop1) {Dropout(0.3)};

\node[convblock, right=0.6cm of conv1] (conv2) {Conv1D\\128, k=3};
\node[box, fill=blue!8, below=0.08cm of conv2] (bn2) {BatchNorm};
\node[box, fill=blue!8, below=0.08cm of bn2] (pool2) {MaxPool(2)};
\node[box, fill=blue!8, below=0.08cm of pool2] (drop2) {Dropout(0.3)};

\node[lstmblock, right=0.6cm of conv2, minimum width=2.3cm] (bilstm) {Bidirectional\\LSTM (64$\times$2)};
\node[box, fill=green!8, below=0.08cm of bilstm, minimum width=2.3cm] (drop3) {Dropout(0.3)};

\node[denseblock, right=0.6cm of bilstm] (dense1) {Dense(128)\\+ ReLU};
\node[box, fill=orange!8, below=0.08cm of dense1] (bn3) {BatchNorm};
\node[box, fill=orange!8, below=0.08cm of bn3] (drop4) {Dropout(0.3)};

\node[ioblock, right=0.6cm of dense1] (output) {Softmax\\9 classes};

\draw[arrow] (input) -- (conv1);
\draw[arrow] (conv1) -- (conv2);
\draw[arrow] (conv2) -- (bilstm);
\draw[arrow] (bilstm) -- (dense1);
\draw[arrow] (dense1) -- (output);

\node[dimlab, below=0.2cm of drop1] {$64 \times 64$};
\node[dimlab, below=0.2cm of drop2] {$32 \times 128$};
\node[dimlab, below=0.2cm of drop3] {$128$};

\node[dimlab, above=0.2cm of input, text width=1.8cm, align=center] {40 MFCC + 40 Mel\\+ 12 Chroma};

\begin{scope}[on background layer]
    \node[draw=blue!50, dashed, rounded corners, fit=(conv1)(bn1)(pool1)(drop1), inner sep=3pt, label={[label, text=blue!70]above:CNN Block 1}] {};
    \node[draw=blue!50, dashed, rounded corners, fit=(conv2)(bn2)(pool2)(drop2), inner sep=3pt, label={[label, text=blue!70]above:CNN Block 2}] {};
    \node[draw=green!50, dashed, rounded corners, fit=(bilstm)(drop3), inner sep=3pt, label={[label, text=green!60!black]above:Temporal}] {};
    \node[draw=orange!50, dashed, rounded corners, fit=(dense1)(bn3)(drop4), inner sep=3pt, label={[label, text=orange!70!black]above:Classifier}] {};
\end{scope}

\end{tikzpicture}
}

\caption{CNN-BiLSTM for technique classification. Tensor dimensions shown in gray ($\text{time} \times \text{features}$).}
\label{fig:cnn_bilstm_arch}
\end{figure*}

The classifier itself is a temporal CNN-BiLSTM, illustrated in Figure~\ref{fig:cnn_bilstm_arch}. Two 1D convolutional blocks with 64 and 128 filters (kernel size 3), each followed by batch normalization, max pooling (factor 2), and dropout (0.3), feed into a bidirectional LSTM with 64 units per direction (128 total). A fully-connected classification head (128 units, ReLU, batch normalization, dropout) and a softmax over the nine technique classes produce the final prediction. The model has approximately 160{,}000 parameters in total.

\subsection{Training}
\label{ssec:stage2_training}

A central challenge in expressive technique classification is that no single public dataset covers the full range of techniques used in real guitar performance, and existing datasets adopt incompatible label schemes. We consolidate five publicly available guitar technique datasets into a single unified training set: IDMT-SMT-Chords~\cite{kehling_IDMT}, Guitar-TECHS~\cite{pedroza2025guitartechselectricguitardataset}, AGPT~\cite{Stefani_agpt}, EG-IPT~\cite{fiorini_2025_15205644}, and Magcil~\cite{Mitsou_magcil}. We preprocess each dataset to extract individual technique segments under a common nine-class taxonomy, then apply a stratified 72/8/20 train/validation/test split, ensuring that multiple microphone captures of the same performance never cross split boundaries.

We train with Adam (learning rate $10^{-3}$) for 200 epochs using sparse categorical cross-entropy loss, with per-class weighting inversely proportional to sample frequency to handle class imbalance. We halve the learning rate when validation loss plateaus for 5 epochs and stop early after 15 epochs without improvement on validation accuracy.

\subsection{Evaluation}
\label{ssec:stage2_eval}

\begin{table}
\centering
\small
\setlength{\tabcolsep}{4pt}
\resizebox{\columnwidth}{!}{%
\begin{tabular}{lcccc}
\toprule
\textbf{Class} & \textbf{Precision} & \textbf{Recall} & \textbf{F1} & \textbf{Support} \\
\midrule
Bend & 92.6\% & 95.1\% & 93.8\% & 183 \\
Hammer/Pull-off & 98.3\% & 96.4\% & 97.4\% & 2,343 \\
Harmonics & 92.2\% & 97.7\% & 94.9\% & 617 \\
Kick Drum & 96.7\% & 99.5\% & 98.1\% & 441 \\
Palm Muting & 98.8\% & 97.0\% & 97.9\% & 1,952 \\
Picking/No Technique & 98.7\% & 97.3\% & 98.0\% & 3,919 \\
Slide & 90.4\% & 99.3\% & 94.7\% & 410 \\
Snare Drum & 99.1\% & 99.5\% & 99.3\% & 1,265 \\
Vibrato & 83.5\% & 96.3\% & 89.4\% & 242 \\
\midrule
\textbf{Macro Avg} & 94.5\% & 97.6\% & 95.9\% & 11,372 \\
\textbf{Weighted Avg} & 97.5\% & 97.4\% & 97.4\% & 11,372 \\
\bottomrule
\end{tabular}%
}
\caption{Per-class results on the unified dataset test set.}
\label{tab:classification_results}
\end{table}

Table~\ref{tab:classification_results} reports per-class performance on the unseen test split. All classes exceed 95\% recall, with percussion-style techniques (kick drum, snare drum) reaching 99.5\%. The hardest class is vibrato (89.4\% F1), which is primarily confused with bend due to their shared characteristics. To contextualize these numbers, we compare against two prior methods on the same unified dataset under identical settings: a convolutional network from Fiorini et al.~\cite{fiorini_2025_15205644} and a dense MLP from Stefani et al.~\cite{Stefani_ExpressiveGuitar-TechniqueClassification}, both reimplemented as baselines. As shown in Table~\ref{tab:comparison_prior_work}, our CNN-BiLSTM substantially outperforms both (95.9\% Macro F1 vs.\ 71.6\% and 62.7\%) with a 13$\times$ parameter reduction relative to Stefani et al.'s MLP, suggesting that explicitly modeling temporal dependencies is more effective and parameter-efficient than scaling feedforward capacity.

\begin{table}
\centering
\small
\begin{tabular}{lccc}
\toprule
\textbf{Model} & \textbf{Params} & \textbf{Accuracy} & \textbf{Macro F1} \\
\midrule
Fiorini et al.~\cite{fiorini_2025_15205644} & 3.70M & 64.3\% & 62.7\% \\
Stefani et al.~\cite{Stefani_ExpressiveGuitar-TechniqueClassification}  & 2.08M & 86.7\% & 71.6\% \\
\textbf{TART (Ours)} & \textbf{160K} & \textbf{97.4\%} & \textbf{95.9\%} \\
\bottomrule
\end{tabular}
\caption{Comparison with prior models on the unified dataset. All models use identical splits.}
\label{tab:comparison_prior_work}
\end{table}

\section{Third Stage: String and Fret Assignment}
\label{sec:string_fret}

The guitar is pitch-redundant, meaning the same MIDI pitch $p_i$ can be produced at multiple different string-fret combinations $(s, f) \in \mathcal{S} \times \mathcal{F}$, where $\mathcal{S} = \{1, \ldots, 6\}$ and $\mathcal{F} = \{0, \ldots, 24\}$. Given Stage~1's note events $\mathcal{E} = \{(p_i, t_i^{\mathrm{on}}, t_i^{\mathrm{off}}, v_i)\}_{i=1}^{|\mathcal{E}|}$ and the input audio $x \in \mathbb{R}^{N}$, Stage~3 assigns each note a string-fret pair $(s_i, f_i)$, producing
\begin{equation}
\label{eq:stage3_output}
\mathcal{E}^{sf} = \{(p_i, t_i^{\mathrm{on}}, t_i^{\mathrm{off}}, v_i, s_i, f_i)\}_{i=1}^{|\mathcal{E}|}.
\end{equation}
Because simple playability heuristics may not align with the performer's ground-truth fingering choices, we infer performer-consistent fingering patterns from both musical context and audio timbral cues.

\subsection{Architecture}
\label{ssec:stage3_arch}

\begin{figure}[!t]
\centering
\resizebox{1.0\columnwidth}{!}{%
\begin{tikzpicture}[
    node distance=0.25cm,
    box/.style={rectangle, draw, minimum width=2.1cm, minimum height=0.5cm, align=center, font=\footnotesize, inner sep=2pt},
    convblock/.style={box, fill=blue!15},
    subblock/.style={box, fill=blue!8, font=\scriptsize, minimum height=0.42cm, minimum width=2.1cm},
    symblock/.style={box, fill=purple!15},
    fuseblock/.style={box, fill=teal!15},
    t5block/.style={box, fill=green!15},
    outblock/.style={box, fill=orange!15},
    ioblock/.style={box, fill=gray!15, rounded corners},
    pitchblock/.style={box, fill=red!12},
    arrow/.style={-{Stealth[length=1.6mm, width=1.1mm]}, thick},
    dasharrow/.style={-{Stealth[length=1.6mm, width=1.1mm]}, thick, dashed, gray},
    dimlab/.style={font=\tiny, text=gray},
    label/.style={font=\scriptsize, align=center}
]

\node[ioblock] (audio_in) {Per-note audio\\200\,ms};
\node[convblock, above=0.35cm of audio_in] (mel) {Log-Mel\\$n_\text{mels}{=}64$};

\node[convblock, above=0.4cm of mel] (conv1) {Conv2D, 32 ch, k=3};
\node[subblock, above=0.08cm of conv1] (bn1) {BN + ReLU};
\node[subblock, above=0.08cm of bn1] (pool1) {MaxPool(2)};

\node[convblock, above=0.3cm of pool1] (conv2) {Conv2D, 64 ch, k=3};
\node[subblock, above=0.08cm of conv2] (bn2) {BN + ReLU};
\node[subblock, above=0.08cm of bn2] (pool2) {MaxPool(2)};

\node[convblock, above=0.3cm of pool2] (conv3) {Conv2D, 128 ch, k=3};
\node[subblock, above=0.08cm of conv3] (bn3) {BN + ReLU};
\node[subblock, above=0.08cm of bn3] (pool3) {MaxPool(2)};

\node[convblock, above=0.3cm of pool3] (flat) {Flatten};
\node[subblock, above=0.08cm of flat] (ln) {Linear$\to$256 + LayerNorm};

\node[ioblock, right=2.3cm of audio_in] (midi_in) {Stage~1 MIDI\\(NOTE\_ON/OFF)};
\node[symblock, above=0.35cm of midi_in] (tokens) {Tokenize\\\texttt{NOTE\_ON<p>}};
\node[symblock, above=0.4cm of tokens] (tokemb) {Token\\Embedding};
\node[pitchblock, above=0.4cm of tokemb] (pitchemb) {Pitch Embed\\$\mathbb{R}^{128{\to}256}$};

\node[fuseblock, above=0.7cm of ln, xshift=1.9cm, minimum width=2.8cm] (fuse) {Concat + Linear\\(512$\to$256) + ReLU};
\node[fuseblock, above=0.3cm of fuse, minimum width=2.8cm] (prefix) {Prepend audio tokens before MIDI sequence};

\node[t5block, above=0.4cm of prefix, minimum width=5.0cm, minimum height=1.0cm] (t5) {T5 Encoder-Decoder\\6 layers, 8 heads, $d{=}256$\\gated-GELU};

\node[outblock, above=0.4cm of t5, minimum width=5.0cm] (constr) {Constrained Beam Search\\(beam=4, chord feas., $\leq 5$-fret span)};

\node[ioblock, above=0.35cm of constr, minimum width=5.0cm] (output) {Output: \texttt{TAB<s,f>\ \ TIME\_SHIFT<d>\ \ $\dots$}};

\draw[arrow] (audio_in) -- (mel);
\draw[arrow] (mel) -- (conv1);
\draw[arrow] (pool1) -- (conv2);
\draw[arrow] (pool2) -- (conv3);
\draw[arrow] (pool3) -- (flat);

\draw[arrow] (midi_in) -- (tokens);
\draw[arrow] (tokens) -- (tokemb);
\draw[dasharrow] (tokens.east) -- ++(0.25,0) |- (pitchemb.east);

\draw[arrow] (ln.north) |- (fuse.west);
\draw[arrow] (pitchemb.north) |- (fuse.east);
\draw[arrow] (tokemb.east) -- ++(0.4,0) |- (prefix.east);

\draw[arrow] (fuse) -- (prefix);
\draw[arrow] (prefix) -- (t5);
\draw[arrow] (t5) -- (constr);
\draw[arrow] (constr) -- (output);

\begin{scope}[on background layer]
    \node[draw=blue!50, dashed, rounded corners,
          fit=(mel)(conv1)(bn1)(pool1)(conv2)(bn2)(pool2)(conv3)(bn3)(pool3)(flat)(ln),
          inner sep=4pt,
          label={[label, text=blue!70]left:Audio\\Encoder\\(CNN)}] {};
    \node[draw=purple!50, dashed, rounded corners,
          fit=(tokens)(tokemb),
          inner sep=4pt,
          label={[label, text=purple!70]right:Symbolic\\Path}] {};
    \node[draw=teal!50, dashed, rounded corners,
          fit=(fuse)(prefix),
          inner sep=4pt,
          label={[label, text=teal!70!black]left:Fusion}] {};
\end{scope}

\end{tikzpicture}
}
\caption{AudioFret architecture for string-fret assignment.}
\label{fig:audiofret}
\end{figure}
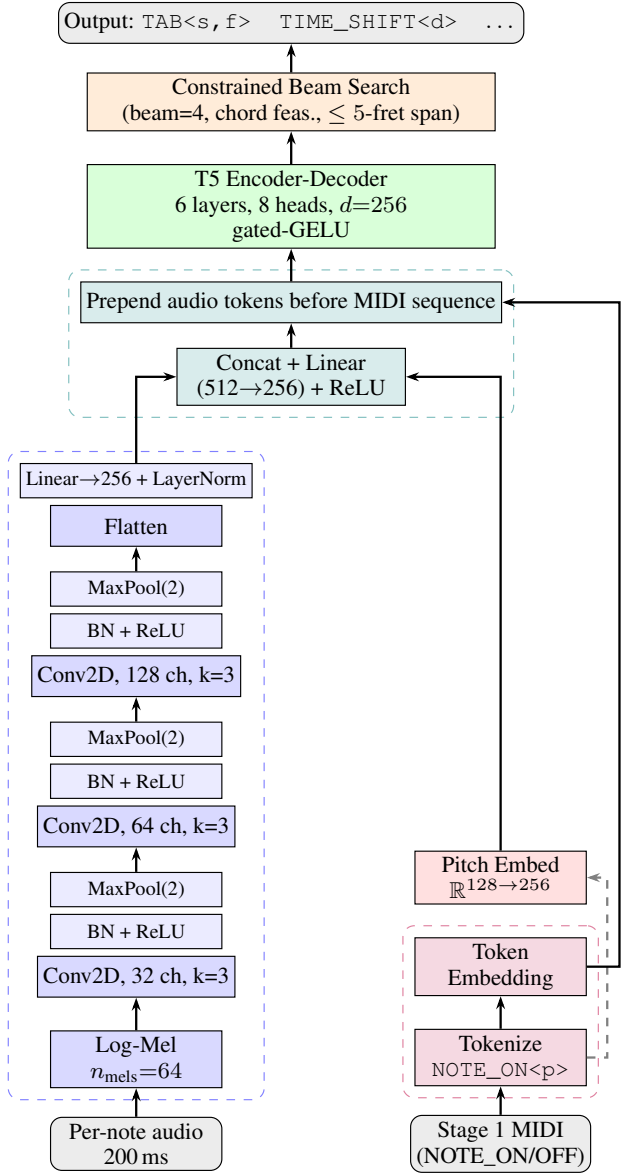

We build upon the Fretting-Transformer~\cite{hamberger2025frettingtransformerencoderdecodermodelmidi}, which treats string and fret assignment as sequence-to-sequence translation using a T5-style encoder-decoder architecture~\cite{raffel2020exploring} with a unified vocabulary for MIDI and tablature tokens. We extend this architecture to produce a new model, which we call \textbf{AudioFret} (Figure~\ref{fig:audiofret}). First, we scale the backbone from the original smaller configuration ($d_{\text{model}}\!=\!128$, 3 layers) to a larger variant ($d_{\text{model}}\!=\!256$, 6 layers, 8 heads, $\sim$15\,M parameters) with gated-GELU feed-forward layers~\cite{shazeer2020glu}, which we find improves cross-dataset generalization. Second, per-note timbral features from the raw audio are injected into the encoder alongside the symbolic MIDI tokens. This directly addresses the pitch-redundancy problem by allowing the model to exploit the fact that the same pitch played on different strings produces audibly different spectra due to differences in string gauge, tension, and overtone structure.

For each input note we extract a 200\,ms mel-spectrogram around its onset, pass it through a lightweight CNN (three convolutional blocks with batch normalization and ReLU), and prepend the resulting per-note audio embeddings as a contiguous block immediately before the MIDI token sequence (after the capo and tuning conditioning tokens). The T5-style self-attention then fuses audio and symbolic information end-to-end without explicit fusion hyperparameters. We retain the tokenization scheme and the capo and tuning conditioning tokens from the original Fretting-Transformer.

\subsection{Training}
\label{ssec:stage3_training}

Training proceeds in two phases. In the first phase, the scaled T5-style backbone is pre-trained from scratch on the combined datasets of SynthTab~\cite{Zang_2024} and DadaGP~\cite{sarmento2021dadagpdatasettokenizedguitarpro}, consisting of 17{,}255 training and 1{,}859 validation tracks. To improve robustness, we augment the symbolic data with capo positions 0--7 and four tuning variants (standard, half-step down, full-step down, drop D), represented as conditioning tokens prepended to each sequence. We pre-train with Adafactor (learning rate $1\!\times\!10^{-4}$, batch size 16) for up to 120 epochs, with early stopping on validation accuracy (patience 15). In the second phase, we pool together GAPS~\cite{riley2024gapslargediverseclassical}, GOAT~\cite{loth2025goatlargedatasetpaired} (DI subset only), and Guitar-TECHS~\cite{pedroza2025guitartechselectricguitardataset} and apply an 85/15 train/validation split. We first pretrain a CNN audio encoder as a string classifier on this dataset (40 epochs, AdamW, learning rate $1\!\times\!10^{-3}$, batch size 64). We then jointly fine-tune the CNN audio encoder and the scaled T5-style backbone end-to-end for 30 epochs with AdamW (base learning rate $5\!\times\!10^{-5}$ for the backbone and $2.5\!\times\!10^{-4}$ for the audio encoder, batch size 8, weight decay $0.01$, and early stopping with patience 8).

\subsection{Post-processing}
\label{ssec:stage3_postproc}

At inference time we replace greedy decoding with beam search (beam width 4) and apply two constraints at each generation step. First, we enforce the tablature token grammar: every note must be followed by a valid duration token, and within a single chord, no string may be assigned twice. Second, we mask out any chord combination that requires a fret span greater than 5 to play, satisfying the realistic reach limit of a guitarist's hand.
\FloatBarrier

\subsection{Evaluation}
\label{ssec:stage3_eval}

We report Tab~F1 as the primary metric, where a predicted note is counted as correct only if onset, pitch, \emph{and} string assignment all match the ground truth (using the same $\pm 50$\,ms onset / $\pm 50$\,cents pitch tolerances as in Stage~1). All results are reported in a zero-shot setting on GuitarSet and EGDB, including their noisy counterparts. Stage~3 is evaluated in an \textbf{oracle} setting, where the model receives ground-truth MIDI events along with the corresponding audio to make predictions. We compare \textbf{AudioFret} against \textbf{TabCNN}~\cite{wiggins2019tabcnn}, the original \textbf{Fretting-Transformer}~\cite{hamberger2025frettingtransformerencoderdecodermodelmidi}, and two ablations of our model: a \textbf{Symbolic-only} variant (our scaled T5 backbone with no audio conditioning) and an \textbf{Audio-only} variant (the CNN string classifier with no symbolic decoder).

\begin{table}
\centering
\small
\setlength{\tabcolsep}{4pt}
\renewcommand{\arraystretch}{1.15}
\resizebox{\columnwidth}{!}{%
\begin{tabular}{@{}lccccc@{}}
\hline
\textbf{Model} & \textbf{GS} & \textbf{GS noi} & \textbf{EGDB} & \textbf{EGDB noi} & \textbf{Avg} \\
\hline
TabCNN~\cite{wiggins2019tabcnn}     & -- & -- & 30.4 & 25.4 & 27.9 \\
Fretting-Transformer~\cite{hamberger2025frettingtransformerencoderdecodermodelmidi}          & 60.4 & 60.4 & 66.3 & 66.3 & 63.3 \\
Audio-only (ours)             & 54.8 & 55.2 & 64.8 & 58.9 & 58.4 \\
Symbolic-only (ours)          & 61.3 & 61.3 & 72.7 & 72.7 & 67.0 \\
\textbf{AudioFret (ours)}     & \textbf{69.2} & \textbf{69.5} & 74.8 & \textbf{73.7} & \textbf{71.8} \\
\hline
\end{tabular}
}
\caption{Zero-shot note-level Tab~F1 (\%) on GuitarSet, EGDB, Noisy GuitarSet, and Noisy EGDB. Since TabCNN was trained on GuitarSet, we report only its EGDB results.}
\label{tab:stage3-eval}
\end{table}

\begin{figure}[t]
    \centering
    \includegraphics[width=1.0\columnwidth]{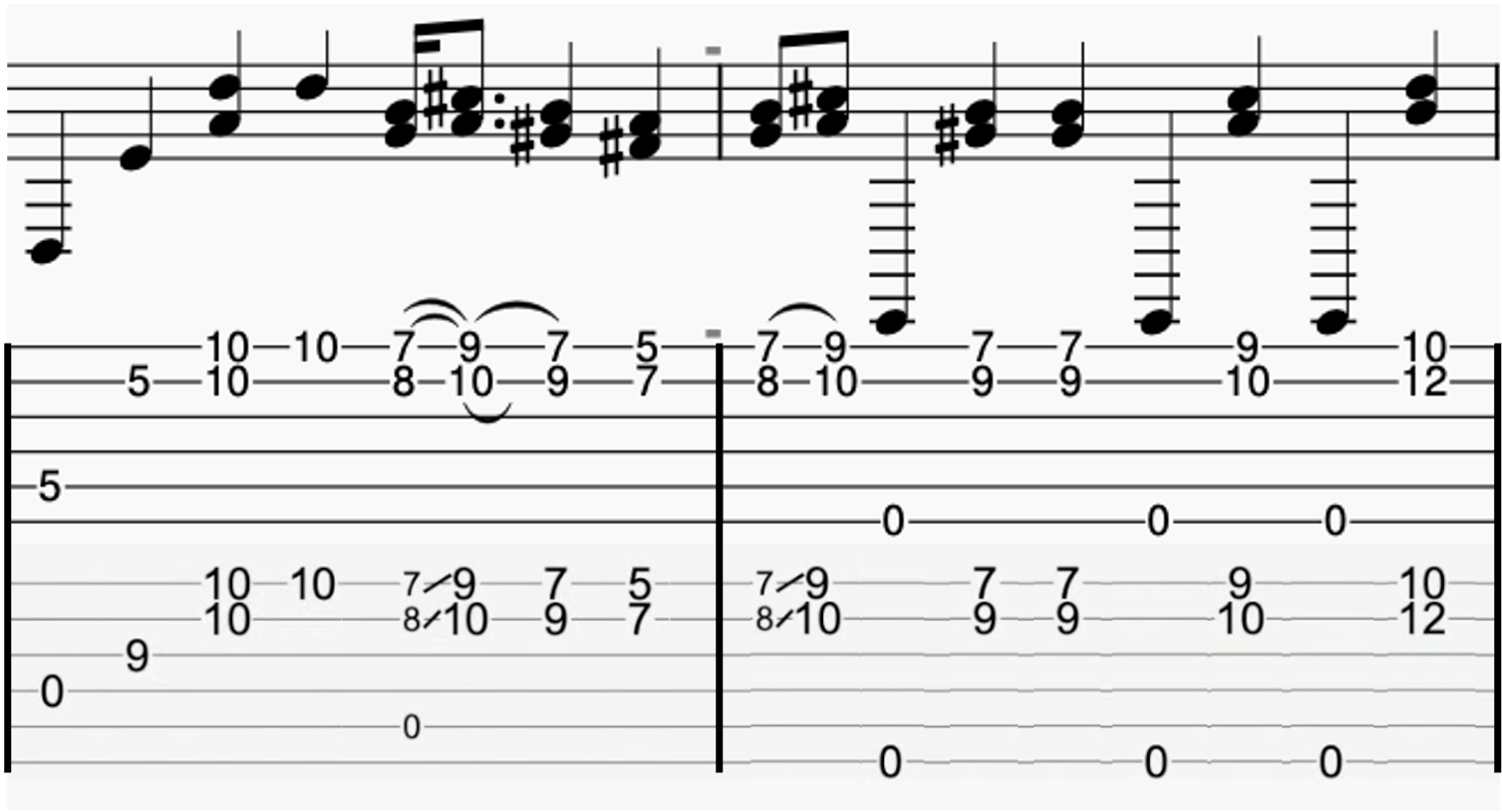}
    \caption{A sample of the chorus of ``Tears In Heaven'' by Eric Clapton, played by YouTube guitarist Kenneth Acoustic (bottom) and the transcription by TART (top). TART occasionally confuses slides with hammer-ons, but otherwise closely matches the ground-truth tablature.}
    \label{fig:tih}
\end{figure}

As shown in Table~\ref{tab:stage3-eval}, scaling the symbolic backbone from the original Fretting-Transformer raises average Tab~F1 from $63.3\%$ to $67.0\%$, a gain of $+3.7$ percentage points without introducing any audio information. Adding audio conditioning on top of the scaled backbone yields a further $+4.8$ percentage points, for a total improvement of $+8.5$ percentage points over the original Fretting-Transformer baseline. AudioFret achieves the best overall Tab~F1 of \textbf{71.8\%}, along with a pitch accuracy of $100.0\%$.

\section{Fourth Stage: Tablature Generation}
\label{sec:illust}

Stage~4 merges the parallel outputs of Stages~2 and~3 (the technique-annotated events $\mathcal{E}^{\tau}$ and the string-fret assignments $\mathcal{E}^{sf}$) together with a tempo estimate $\beta$ from BeatNet~\cite{heydari_beatnet_2021}, into a single fully-annotated event stream stored in a JAMS annotation file:
\begin{equation}
\label{eq:stage4_output}
\mathcal{E}^{\text{final}} = \{(p_i, t_i^{\mathrm{on}}, t_i^{\mathrm{off}}, v_i, \tau_i, s_i, f_i)\}_{i=1}^{|\mathcal{E}|}.
\end{equation}
Stage~4 then renders the pair $(\mathcal{E}^{\text{final}}, \beta)$ into a beat-aligned MusicXML tablature score.

The first step for conversion to MusicXML is rhythmic quantization. We compile a quantized event series from $\mathcal{E}^{\text{final}}$ by clustering onsets within 30\,ms to remove micro-timing errors and quantizing both onsets and durations to a 1/16-note grid using BeatNet's estimated tempo. Notes sharing the same quantized onset are grouped into chords. Same-string collisions are resolved by retaining the longer-duration note.

The second step is score rendering. We export the quantized event stream to MusicXML. Technique annotations from the JAMS file that pertain to a single note (bend, harmonic, vibrato, palm muting, kick drum, snare drum, and picking) are written directly, while techniques that pertain to two consecutive notes (hammer-on/pull-off and slide) are inferred by pairing the note with the next note on the same string within a 1-beat window. Users may optionally provide information on capo position, tuning, and tempo if known; otherwise standard tuning and no capo are assumed. Figure~\ref{fig:finalpipelineflowchart} shows the full TART pipeline, and Figure~\ref{fig:tih} compares tablature generated from TART to ground truth on a section of the song ``Tears in Heaven.''

\section{End-To-End Evaluation}
\label{sec:evaluation}

\begin{table}[t]
\centering
\small
\setlength{\tabcolsep}{4pt}
\resizebox{\columnwidth}{!}{%
\begin{tabular}{lccccc}
\toprule
\textbf{Setting} & \textbf{GS} & \textbf{GS noi} & \textbf{EGDB} & \textbf{EGDB noi} & \textbf{Avg} \\
\midrule
Oracle Tab F1          & 69.2 & 69.5 & 74.8 & 73.7 & 71.83 \\
End-to-end Tab F1      & 56.0 & 51.2 & 55.2 & 53.9 & 54.08 \\
Propagation Cost ($\Delta$) & 13.2 & 18.4 & 19.7 & 19.8 & 17.75 \\
\bottomrule
\end{tabular}
}
\caption{Error propagation for AudioFret.}
\label{tab:error-prop}
\end{table}

TART's pipeline is structured in a way such that technique detection and string-fret predictions require Stage~1's MIDI output as input. Therefore, Stage~1 errors directly impact Stage~2's and Stage~3's end-to-end performance. To measure this effect, we compare AudioFret results using two separate metrics:
\begin{itemize}[leftmargin=*, topsep=4pt, itemsep=3pt, parsep=0pt]
    \item \textbf{Oracle Tab F1.} AudioFret is given the ground-truth MIDI as input (same as Stage~3 results).
    \item \textbf{End-to-end Tab F1.} AudioFret is given Stage~1's predicted MIDI as input.
\end{itemize}

The difference between the two, which we call the \textbf{propagation cost}, is the Tab F1 that the system loses because Stage~1 is imperfect.

\label{sec:end2end}
As shown in Table~\ref{tab:error-prop}, the full pipeline achieves an end-to-end Tab F1 score of \textbf{54.08\%} across the four zero-shot test datasets, with consistent performance on both clean and noisy audio. The drop in overall performance due to propagation error between Stage 1 and Stage 3 is \textbf{17.75} percentage points.

\section{Conclusion}
\label{sec:conclusion}

We present \textbf{TART}, a four-stage pipeline that converts guitar audio into tablature with both expressive technique labels and playable string-fret fingerings. TART combines a noise-robust audio-to-MIDI transcriber, a temporal technique classifier trained on a newly unified nine-class dataset, and AudioFret, a novel audio-conditioned T5 model that exploits timbral cues that purely symbolic models cannot access. Across four zero-shot benchmarks, TART outperforms prior systems.

Several limitations remain. The audio-to-MIDI stage does not detect unpitched notes such as percussion, which prevents downstream technique annotation. In addition, the expressive technique classification stage only assigns a single technique per note, failing to capture simultaneous techniques (e.g., a bend performed with vibrato). Furthermore, tablature generation relies on fixed-rhythmic quantization for chord grouping, which can produce occasional misgroupings. Nevertheless, TART closes a long-standing gap in guitar AMT by jointly modeling pitch, expressive technique, and string-fret position in generated tablature.

\bibliography{ISMIRtemplate}

%
%
%
%

\end{document}